\documentclass{article}
\usepackage{microtype}
\usepackage{graphicx}
\usepackage{subfigure}
\usepackage{tabularx}
\usepackage{booktabs} 
\usepackage{amsmath} 
\usepackage{enumitem}
\usepackage{amssymb}
\usepackage{hyperref}
\usepackage{multirow}
\usepackage[table]{xcolor}

\usepackage[accepted]{mlsys2025}

\mlsystitlerunning{Submission and Formatting Instructions for MLSys 2025}

\begin{document}
\twocolumn[
\mlsystitle{Jenius Agent: Towards Experience-Driven Accuracy Optimization in Real-World Scenarios}







\begin{mlsysauthorlist}

\mlsysauthor{Defei Xia}{tjh}  
\mlsysauthor{Bingfeng Pi}{tjh}
\mlsysauthor{Shenbin Zhang}{tjh}
\mlsysauthor{Song Hua}{tjh}
\mlsysauthor{Yunfei Wei}{tjh}
\mlsysauthor{Lei Zuo}{tjh}
\end{mlsysauthorlist}
\mlsysaffiliation{tjh}{Tianju Dihe (Suzhou) Technology Co., Ltd., Suzhou, China}
\mlsyscorrespondingauthor{Defei Xia}{xiadf@think-land.com}

\mlsyskeywords{Autonomous Agent; Large Language Model; Context
Understanding; Tool Usage}

\vskip 0.3in

\begin{abstract}
As agent systems powered by large language models (LLMs) advance, improving performance in context understanding, tool usage, and long-horizon execution has become critical. However, existing agent frameworks and benchmarks provide limited visibility into execution-level behavior, making failures in tool invocation, state tracking, and context management difficult to diagnose. This paper presents Jenius-Agent, a system-level agent framework grounded in real-world deployment experience. It integrates adaptive prompt generation, context-aware tool orchestration, and layered memory mechanism to stabilize execution and improve robustness in long-horizon, tool-augmented tasks. Beyond system design, we introduce an evaluation methodology that jointly measures procedural fidelity, semantic correctness, and efficiency. This framework makes agent behavior observable as a structured execution process and enables systematic analysis of failure modes not captured by output-only metrics. Experiments on Jenius-bench show substantial improvements in task completion rate, with up to a 35\% relative gain over the base agent, along with reduced token consumption, response latency, and tool invocation failures. The framework is already deployed in Jenius (\url{https://www.jenius.cn}), providing a lightweight and scalable solution for robust, protocol-compatible autonomous agents.
\end{abstract}
]



\printAffiliationsAndNotice{}  

\section{Introduction}
In the context of the increasing capability of large language models (LLMs), LLM-based autonomous agents have become a new paradigm for AI applications~\cite{yao2023react,xi2025rise, wang2024survey}. These agents are capable of understanding instructions, invoking tools, reasoning and planning, and performing complex tasks, and are widely used in a variety of fields such as research assistants, process automation, retrieval-augmented generation, and code generation and debugging~\cite{grattafiori2024llama,masterman2024landscape,qin2024tool,singh2025agentic}. Although current intelligent agent systems (e.g., AutoGPT, LangChain Agents, BabyAGI) ~\cite{yang2023auto,topsakal2023creating,nakajima2023task} have begun to take shape, there are still numerous challenges in their generality, stability, and manageability, especially in task precision, response reliability, and system stability in situations with many tasks~\cite{cemri2025multi}. 

Most existing agent systems rely on fixed prompts and predefined tool-use workflows, limiting their ability to understand task intent, select tools dynamically, and manage context effectively~\cite{e2025rag,qin2024tool,schick2023toolformer}. Numerous works have shown that dynamic prompt, tool retrieval, and memory management play a key role in optimizing agent execution. For instance, MCP-Zero~\cite{fei2025mcp} enables active tool discovery for unseen tasks, while multi-agent designs~\cite{zhou2025multi} enhance planning and coordination through optimized prompts and collaboration strategies. The Model Context Protocol (MCP)~\cite{hou2025model} standardizes context exchange, ensuring reliable tool invocation across systems. Together, these advances underscore that robust prompting, intelligent tool access, and effective memory are essential for high-quality agent execution.

Despite recent advances, autonomous agent pipelines still face three key challenges. First, fixed or generic prompts often misinterpret user intent and fail to adapt to evolving task states, causing unstable behavior and inconsistent outputs. Second, static tool lists or handcrafted rules cannot reliably choose the right tools under ambiguity or across domains, leading to unnecessary or incorrect calls. Finally, long dialogues accumulate redundant context, increasing token cost and diluting salient signals, which weakens reasoning quality.

Beyond system design, evaluating long-horizon, tool-augmented agents remains underdeveloped. Existing benchmarks emphasize final outputs or single-step tool calls. They fail to capture execution failures such as missing tools, incorrect ordering, or partial completion.
This gap motivates an execution-centric evaluation perspective.

This study starts from the basic execution process of autonomous agents and introduces three complementary optimization modules addressing the above issues:

\begin{itemize}[noitemsep, topsep=0pt, leftmargin=*]
\item Task Understanding and Prompt Optimization: Structured intent recognition was combined with refined system prompts and templates to adapt instructions to current state and goals, reducing misinterpretation and stabilizing task alignment.
\item Tool Retrieval: Dynamic retrieval and adaptive tool access were used to match user intent with context-relevant tools and to handle ambiguous user requests.
\item Hierarchical Memory Management: Redundant dialogue history was pruned to control token length, preserving essential semantics and stabilizing reasoning in long-horizon interactions.
\end{itemize}

Rather than evaluating isolated modules, this paper builds a unified framework named Jenius, which improves task accuracy, efficiency, and contextual robustness, aligned with emerging agent communication protocols (e.g., MCP, ACP, A2A)~\cite{ibm2024acp,google2024a2a}. 

The main contributions are summarized as follows:

\begin{itemize}[noitemsep, topsep=0pt, leftmargin=*]
\item We introduced a system-level execution abstraction that models agents as structured processes with explicit state, tool interactions, and failure modes.
\item We proposed a modular optimization framework integrating adaptive prompting, context-aware tool orchestration, and hierarchical memory management to mitigate context noise, tool misuse, and harmful prompting;
\item We improved task grounding and execution robustness through adaptive prompt generation and context-aware tool usage, while introducing preliminary defense strategies to strengthen the system’s resilience to malicious or erroneous inputs;
\item We designed a comprehensive evaluation framework that jointly measures procedural fidelity (4T), semantic quality (CRCFF), and efficiency, enabling diagnosis of real-world agent failures;
\item We also conducted extensive experiments both public and real-world datasets, demonstrating consistent gains in task accuracy, response quality, token efficiency.

\end{itemize}

\begin{figure}[t]
\centering
\includegraphics[scale=0.65]{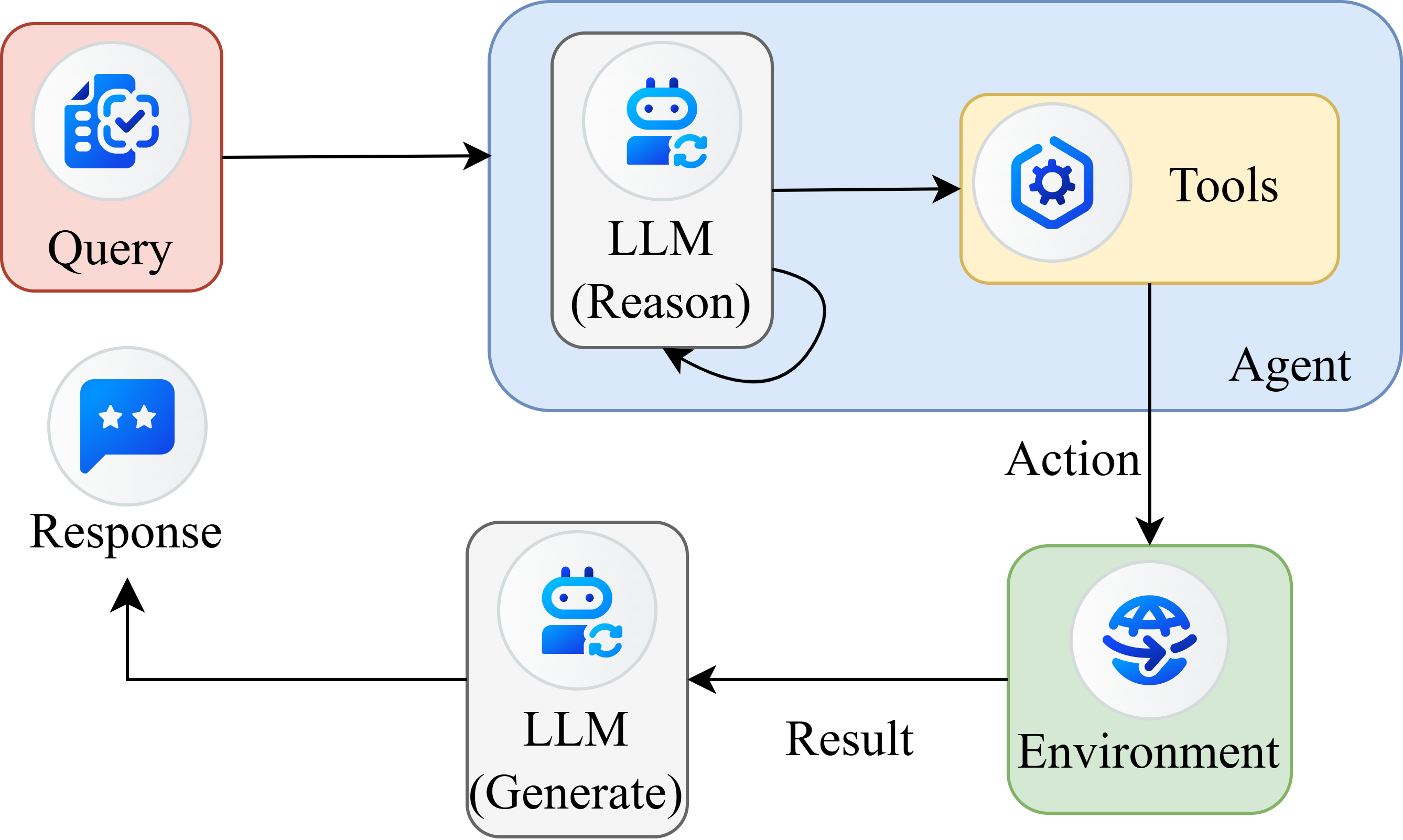}
\caption{A typical ReAct style autonomous agent workflow.}
\label{fig:base_agent}
\end{figure}

\section{Related Work}
With the rapid advancement of LLMs, research on agent systems has expanded in multiple directions. A typical ReAct~\cite{yao2022react} style autonomous agent workflow, illustrated in Figure~\ref{fig:base_agent}, follows an interleaved cycle of reasoning, action, and feedback. In this paradigm, the LLM is no longer a passive text generator but serves as a central planner. This iterative planner–tool–environment interaction highlights three optimization focuses:(i) prompt optimization for reliable reasoning, (ii) tool selection for precise external interactions, and (iii) memory management to preserve context and limit token usage.

\subsection{Prompt Engineering and Optimization}
DSPy~\cite{khattab2024dspy} provides a modular framework for prompt compilation, improving clarity and debuggability, while Prompt Engineering in LLMs~\cite{marvin2023prompt} summarizes practical design principles. Reflect-Retry-Reward~\cite{bensal2025reflect} further introduces adaptive optimization through self-reflection and reinforcement learning. However, current approaches remain unreliable in tool-based or long-horizon tasks, as they often depend on static templates, lack systematic evaluation, and struggle to control structured output or tool invocation consistency.

\subsection{Tool Selection and Invocation}
Recent efforts such as MCP-Zero~\cite{fei2025mcp} explore adaptive tool discovery using semantic embeddings, and BioMedTools~\cite{liu2025biomedtools} illustrates domain-specific coordination via curated tool metadata. However, prior approaches frequently assume clean tool descriptions and fixed invocation formats, whereas real-world systems face noisy metadata, semantic mismatches, and poorly handled runtime errors, resulting in unstable tool retrieval and parameter usage.

\begin{figure*}[t]
\centering
\includegraphics[width=\textwidth]{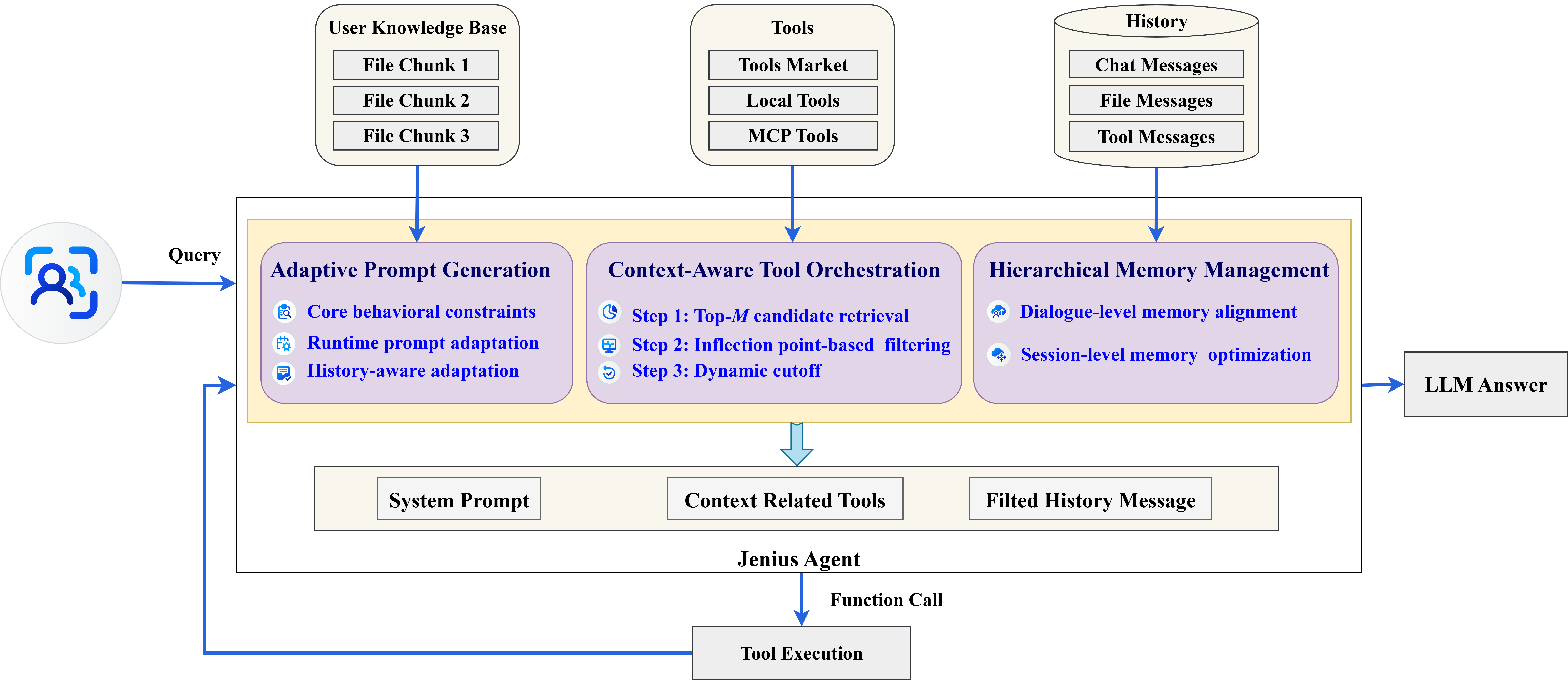}
\caption{Jenius Agent Framework. The LLM acts as the central orchestrator, coordinating task execution with three core modules: adaptive prompt generation, tool retrieval, and memory management that enhance adaptability and efficiency.}
\label{fig:System_F}
\end{figure*}

\subsection{Memory Management}
StateFlow~\cite{wu2024stateflow} and ReAct~\cite{yao2023react} demonstrate state-aware and reasoning-aligned memory design, while Recursively Summarizing~\cite{wang2025recursively} introduces hierarchical compression for long-term coherence.Existing methods often use simple window-based truncation or stage-level summaries, risking loss of critical early context or fine-grained tool dependencies, lacking tight integration between reasoning and memory retention.

Prior agent frameworks differ substantially in execution assumptions. For example, AutoGen~\cite{wu2024autogen} relies on multi-agent message passing with implicit tool semantics, while ReAct assumes a fixed reasoning–action loop. Memory persistence, tool invocation contracts, and prompt adaptation strategies vary across systems. As a result, direct performance comparison often reflects framework assumptions rather than system effectiveness. Our evaluation therefore focuses on controlled, within-framework evolution to isolate the impact of execution-level design choices.

\section{Method} 
To address the above limitations, we proposed a feedback-driven Jenius-agent(Figure~\ref{fig:System_F}), which integrates three coordinated optimization modules. First, adaptive prompt generation creates system prompts by combining role instructions, task state, and user context. Second, tool retrieval selects relevant tools from heterogeneous sources. Third, memory management summarizes historical interactions into compact, semantically enriched representations. 

The LLM produces reasoning traces and function calls, executed by the tool execution module, with results are fed back into the agent. This loop architecture enables continuous integration of prompt adaptation, tool utilization, and memory compression, enhancing both robustness and task alignment.

The following subsections detail the design, motivation, and role of each module.

\subsection{Adaptive Prompt Generation}
\label{sec:dynamic-prompt}
\begin{figure}[t]
\centering
\includegraphics[scale=0.4]{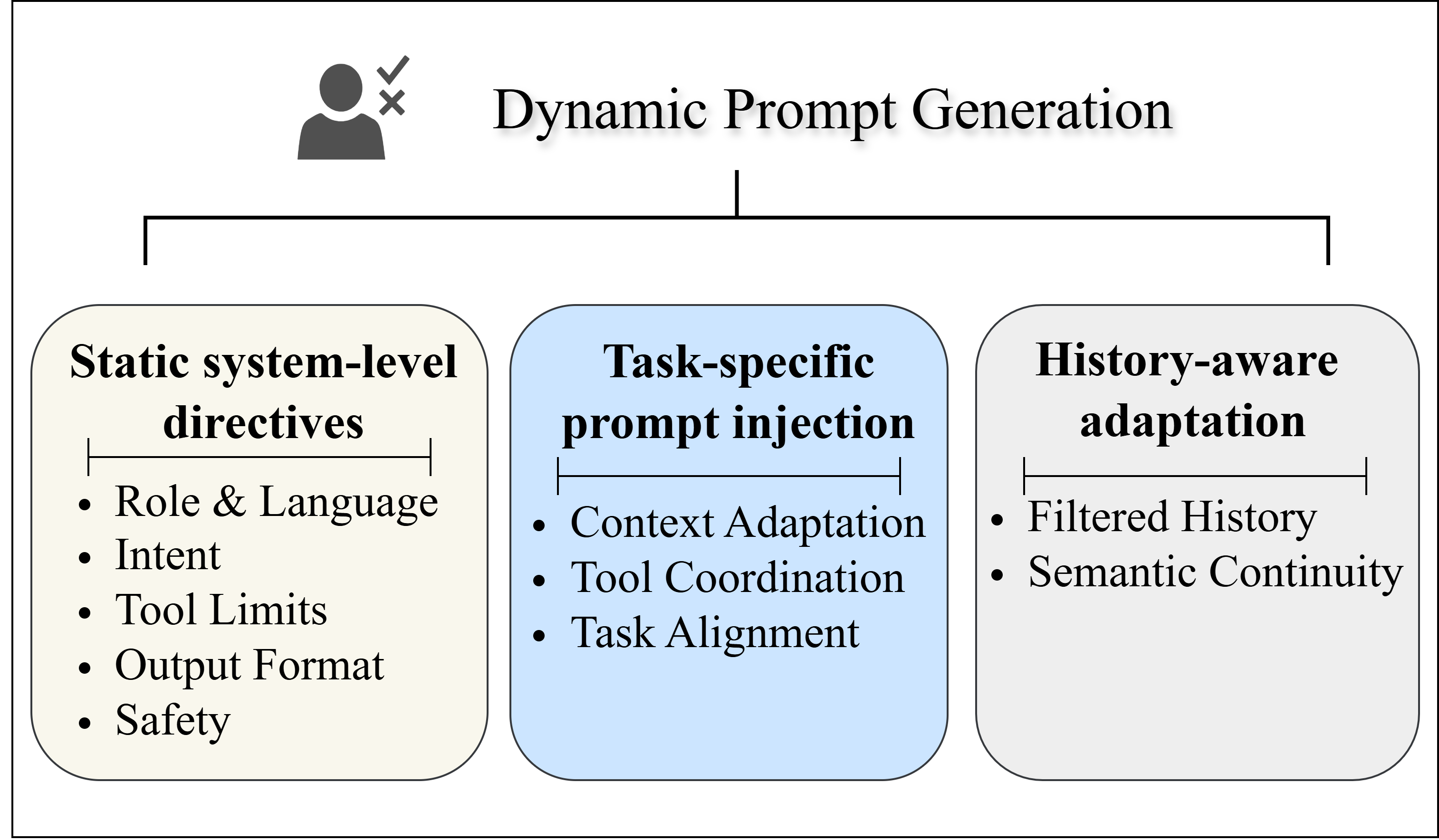}
\caption{Adaptive Prompt Generation.}
\label{fig:DPG}
\end{figure}

Figure~\ref{fig:DPG} illustrates the prompt generation pipeline, which derives adaptive prompts via the fusion of core behavioral constraints on agent, task-driven instructional augmentation, and history-aware adjustments.

\textbf{(1) Core behavioral constraints:} Defines the agent’s role, interaction protocol, response style, and operational limits to ensure consistent, safe, and task-appropriate behavior.

\begin{itemize}[noitemsep, topsep=0pt, leftmargin=*]
\item Role specification and linguistic alignment: Defines the agent’s identity and ensures linguistic coherence by aligning the response language with the user’s input, thereby avoiding unintended multilingual mixing in generated responses; 
\item Intent taxonomy and routing: Classifies user queries into four semantic categories: (i) social interaction (e.g., greetings, small talk), (ii) creative generation (e.g., storytelling, content drafting), (iii) factual recall (queries resolvable via internal knowledge), and
(iv) tool-augmented reasoning, further subdivided into single-tool invocation or multi-step, multi-tool orchestration. This taxonomy enables the dynamic prompt generator to tailor role instructions, reasoning strategies, and output formats according to the detected intent category, thereby ensuring that the generated responses are more precise and contextually appropriate;
\item Controlled tool invocation policy: Implements strict controls over external tool usage to ensure factual accuracy and operational reliability. The agent validates tool eligibility and parameter consistency before invocation, rejecting speculative or unverifiable requests. This mechanism mitigates hallucinations of fabricated entities (e.g., dates, names, locations), prevents redundant or cyclic calls, and enforces deterministic, auditable execution aligned with reliability standards;
\item Response structuring protocol: Determines the output format based on task semantics and usability requirements, such as structured (e.g., tables, JSON, code) or unstructured (free text).
\item Safety and content moderation: Establishes multi-layered guardrails combining hard rule-based filters and soft model-level moderation. These constraints detect and block unsafe, unethical, or privacy-violating outputs while discouraging implicit bias and emotionally manipulative expressions. The moderation layer further curbs hallucinated or contextually inappropriate content that may arise from ambiguous prompts or overgeneralized reasoning. 
\end{itemize}

\textbf{(2) Runtime prompt adaptation:} This component dynamically adapts prompt based on task semantics, tool availability, and interaction context to ensure output alignment with user intent. It follows three principles:
\begin{itemize}[noitemsep, topsep=0pt, leftmargin=*]
\item {Context-aware adaptation:} Prompts are conditioned on the current query, task category, and operational constraints (e.g., tool eligibility, safety rules), guiding appropriate reasoning strategies and response formats.
\item {Tool orchestration guidance:} Conditioning signals direct correct and minimal invocation behavior, suppressing redundant, invalid, or hallucinated calls to improve execution reliability.
\item {Fine-grained intent discrimination:} 
The agent distinguishes between semantically similar but functionally distinct tasks (e.g., textual vs slide report) to ensure fidelity to user intent.
\end{itemize}

By integrating these mechanisms, this component complements the static behavioral constraints, improving task accuracy, reducing erroneous tool usage, and maintaining consistency and relevance across conversational turns.

\textbf{(3) History-aware adaptation:} Rather than relying on raw conversation logs, we employ a relevance-driven compression strategy to filter logs, only preserve key cross-turn dependencies. The resulting history is summarized and appended to the prompt, facilitating precise and coherent action planning. For history filtering details, please refer to section of "Hierarchical Memory Management".

\subsection{Context-Aware Tool Orchestration}
\label{sec:tool-retrieval}
This module is designed to enhance the efficiency of tool access by dynamically selecting appropriate external tools based on the identified task intent. Our framework establishes a structured MCP tool management mechanism, enabling systematic optimization, extensibility, and constraint enforcement over available tools. For instance, third-party MCP tools with inadequate functional descriptions are repackaged and augmented to improve their interpretability and usability. Furthermore, tools are classified according to their functional categories, such as file management, information retrieval, image generation, and data analysis, thereby supporting organized indexing and facilitating efficient, context-aware tool discovery.

All tools are represented as high-dimensional embeddings using the Qwen3 Embedding model~\cite{zhang2025qwen3}, and tool selection is formalized as a semantic similarity ranking process. The similarity between the query and each tool embedding determines their conceptual proximity, which serves as the basis for ranking and subsequent relevance filtering. 

\textbf{Step 1: Top-$\boldsymbol{M}$ candidate retrieval.}  

Candidate tools are ranked by similarity to the query embedding. The top-$M$ tools with the highest similarity values are selected as the initial candidate set, forming a shortlist for further refinement.

\textbf{Step 2: Inflection point-based filtering.}  

To identify a meaningful threshold for relevance, a hybrid inflection point detection method is applied. Two complementary methods are combined: 
\begin{itemize}[noitemsep, topsep=0pt, leftmargin=*]  
\item \textbf{Similarity-jump approach:} Detects the point where cosine similarity values exhibit a sharp decline, marking the transition between relevant and irrelevant tools.  
\item \textbf{Kneedle algorithm:} Detects inflection points in the similarity distribution by analyzing the deviation between the normalized cumulative similarity curve and the diagonal. The point of maximum deviation is chosen as the cutoff threshold for selecting relevant tools.

\end{itemize}


\textbf{Step 3: Dynamic cutoff and final section} 

After both methods, the final retained tool count $N$ is set to the smaller of the two candidate sizes:  

\begin{equation}
N = \min(N_{\text{jump}}, N_{\text{kneedle}}),
\end{equation}
where $N < M$. If $N < 10$, the system supplements the set by selecting the top-10 most similar tools.  

Empirical observations in real-world scenarios showed that overly small $N$ values restricted candidate diversity, whereas excessively large $N$ introduced semantic noise and redundant retrievals. Thus, setting $N=10$ achieved an optimal balance between precision and efficiency.

The refined tool set is passed to the LLM, where vector-based retrieval and inflection-point filtering together ensure accurate, efficient access while minimizing irrelevant calls and maintaining broad task coverage.

Tool orchestration in Jenius is not based solely on similarity ranking. All tools are normalized with explicit schemas, including input constraints, expected outputs, and invocation signatures. Retrieval produces a candidate set, but the final tool choice is resolved during execution based on argument compatibility and runtime feedback.

When tools have overlapping scopes or near-duplicate descriptions, multiple candidates may be retrieved. Ambiguity is resolved at execution time: incompatible tools fail to accept arguments or produce invalid outputs, leading to partial execution rather than silent success. These failures propagate through the execution state and affect subsequent planning steps. Similarly, tools with low-quality or adversarial descriptions are more likely to cause invocation errors or incomplete execution. Such cases increase execution inconsistency and reduce procedural correctness, rather than artificially improving apparent success. This behavior is explicitly exposed at the execution level and analyzed in our evaluation, rather than being masked by output-only correctness signals.

\subsection{Hierarchical Memory Management}
\label{sec:memory-management}
To address the challenge of context expansion in multi-turn dialogues, this module adopts a layered memory mechanism, which integrates fine-grained dialogue-level message alignment and coarse-grained session-level summarization.

\textbf{(1) Dialogue-level memory alignment:} Each round of interaction turn comprises of  a sequence of message types: HumanMessages, AIMessages, and optionally ToolMessages, depending on whether external tools are invoked during the turn. The structure of the dialogue history is preserved as follows:
\begin{itemize}[noitemsep, topsep=0pt, leftmargin=*]
\item \textit{Without tool calls:} The memory typically consists of two messages: one HumanMessage and one corresponding  AIMessage.
\item \textit{With tool calls:} the message flow becomes more complex. If $k$ tools are sequentially invoked in a single turn, the memory sequence contains $2 + 2k$ entries: one HumanMessage, $(k+1)$ AIMessages and $k$ ToolMessages. Concretely, $k=1 \rightarrow 4$ messages (Human, AI, Tool, AI); $k=2 \rightarrow 6$ messages (Human, AI, Tool, AI, Tool, AI); $k=3 \rightarrow 8$ messages, and so on. 
\end{itemize}

To maintain structural consistency and semantic coherence across turns, this module performs pairwise alignment of adjacent messages in the dialogue traces. This alignment process detects and rectifies disruptions in the expected message sequence, particularly those arising from partial or failed tool executions (e.g., user-initiated cancellations, API failures, or LLM output errors) that result in missing or empty ToolMessages. In such cases, the system inserts semantically appropriate placeholder entries or reconstructs the absent information through contextual inference.
Furthermore, to mitigate data loss caused by system faults or token-length limitations during tool invocation, the mechanism employs a backfilling strategy to recover truncated or omitted messages. By reconstructing these missing components based on available context and execution logs, the module ensures the integrity of the canonical interaction pattern: Human → AI (action proposal) → Tool → AI (response synthesis).

Collectively, these operations preserve the temporal and functional continuity of the dialogue state, thereby enhancing robustness and reliability in extended, tool-augmented conversations.

\begin{figure}[t]
\centering
\includegraphics[scale=0.45]{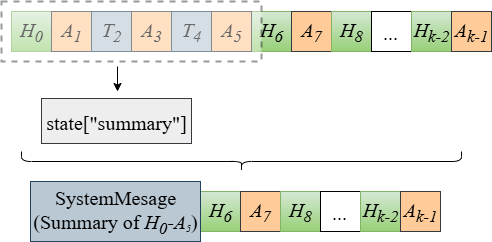}
\caption{Message Summarization Process.}
\label{fig:DMRM}
\end{figure}

\textbf{(2) Session-level memory optimization:} When the cumulative number of messages within a session exceeds a predefined threshold $K$, the system triggers a context summarization mechanism to compress the historical dialogue state, thereby mitigating excessive token consumption and maintaining inference efficiency. Rather than employing naive truncation, which risks loss of critical semantic information, our framework implements a structured summarization strategy designed to preserve conversational coherence and intent continuity.

Specifically, the system identifies the segment for summarization as the subsequence extending from the earliest HumanMessage $H_0$ up to (but excluding) the second-to-last HumanMessage in the current history. This ensures that the most recent interactive context—containing the latest user input and subsequent system responses—remains unsummarized and fully accessible to support ongoing reasoning and tool invocation.
For instance, given a message sequence (as shown in Figure~\ref{fig:DMRM}), when the total message count exceeds capacity $K$, the segment from $H_0$ to $A_5$
is selected for summarization. The resulting summary is stored in the session state under \textit{state["summary"]} and subsequently integrated into the dialogue context as a SystemMessage. This summarized entry replaces the original message block in the history, effectively reducing memory footprint while retaining high-level semantic content.

During model interaction, the summary message is prepended to the active dialogue history, ensuring that salient contextual information is preserved at the beginning of the input sequence. This placement optimizes its retention under attention mechanisms and supports coherent continuation of long-horizon conversations.

By integrating fine-grained dialogue-level alignment with coarse-grained session-level summarization, the proposed memory architecture achieves robust, scalable context management. It effectively balances the trade-off between contextual fidelity and computational efficiency, minimizing redundancy while preserving task-critical intent across extended multi-turn interactions.

\section{Experimental Design and Result Analysis}
We present a evaluation of the proposed optimization modules of adaptive prompt generation, tool orchestration, and memory management, then integrated into several variants of ReAct-style autonomous agent frameworks. The primary objective is to evaluate how these components enhance procedural fidelity and semantic quality in complex, tool-augmented, multi-turn interactions, including token consumption, enabling a comprehensive evaluation of performance and resource utilization in real-world scenarios.

\subsection{Evaluation Metrics and Assessment Criteria}
\label{sec:eva-ver}
The evaluation system consists of the following three parts, as shown in Figure~\ref{fig:evaluation}:

\begin{figure}[t]
\centering
\includegraphics[scale=0.45]{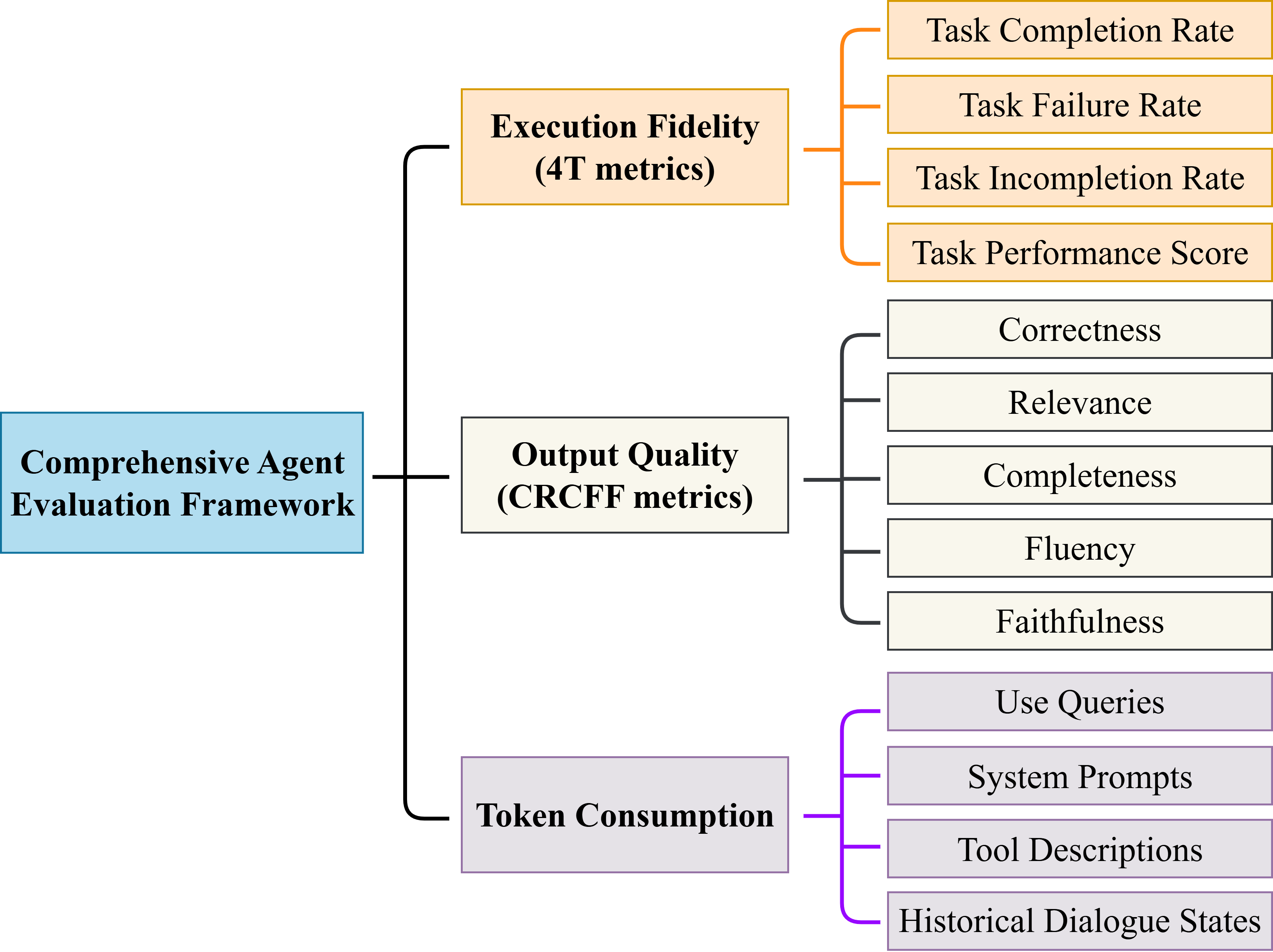}
\caption{Comprehensive agent evaluation framework for Procedural, Semantic, and Efficiency Dimensions.}
\label{fig:evaluation}
\end{figure}

\subsubsection{\textbf{Execution fidelity}}
Examines the agent’s behavioral precision and procedural consistency during task execution, focusing on the sequence of tool invocations and the degree of task completion. 

This assessment goes beyond static output quality by examining the rationality and correctness of the agent’s operational trajectory, focusing on how it is achieved. As such, it better aligns with the functional requirements of autonomous agents in real-world, dynamic environments where correct behavior depends on both accurate decisions and proper execution order.

To quantify execution fidelity, we introduce a structured evaluation framework based on four complementary metrics, collectively termed the \textbf{4T metrics}, to assess critical aspects of tool-mediated reasoning and action:

\begin{itemize}[noitemsep, topsep=0pt, leftmargin=*]
\item \textbf{Task Completion Rate (TCR)}: Proportion of tasks where all required tools are invoked in correct sequence:
\begin{equation}
\text{TCR} = \frac{1}{N} \sum_{i=1}^{N} \mathbb{I}\left[R_i \subseteq L(i)\right], 
\end{equation}
where $N$ is the total number of test tasks, $L(i)$ is the actual tool sequence invoked by the agent on task $i$, and $R_i$ is the human-annotated reference sequence. The indicator function $\mathbb{I}[\cdot]$ returns 1 if $R_i$ is a subsequence of $L(i)$, and 0 otherwise.

\item \textbf{Task Failure Rate (TFR)}: Percentage of tasks with no meaningful execution due to empty tool calls or runtime/system errors:
\begin{equation}
\text{TFR} = \frac{1}{N} \sum_{i=1}^{N} \mathbb{I}\left[L(i) = \emptyset \lor
\text{Error}(i)\right].
\end{equation}
\begin{itemize}[noitemsep, topsep=0pt, leftmargin=*]
  \item No tools were invoked: $L(i) = \emptyset$
  \item Runtime errors (e.g., recursion limit, tool crash)
  \item Agent or system exceptions
\end{itemize}

\item \textbf{Task Incompletion Rate (TIR)}: TIR measures the rate of partially completed tasks due to mistakes in tool selection or sequence, excluding runtime errors. These include missing required tools, misordered invocation, or calling unrelated tools. \begin{equation} \text{TIR} = \frac{1}{N} \sum_{i=1}^{N} \mathbb{I} \left[ R_i \not\subseteq L(i) \land L(i) \neq \emptyset \right]. \end{equation}

\item \textbf{Task Performance Score (TPS)}: Fine-grained measure considering correct, wrong, and missing tools:
\begin{equation}
\begin{gathered}
\text{TPS}(i) = \frac{C_i}{C_i + \lambda_w W_i + \lambda_m M_i}, \\
\text{TPS}_{\text{avg}} = \frac{1}{N}\sum_{i=1}^{N}\text{TPS}(i),
\end{gathered}
\end{equation}

where $C_i$, $W_i$, and $M_i$ denote the number of correctly used, wrongly invoked, and missed tools, respectively. The penalty weights $\lambda_w$ and $\lambda_m$ default to 1 unless otherwise specified.
\end{itemize}

TCR, TFR, and TIR form a complete and mutually exclusive partition of execution outcomes. Their values sum to one for each task.
This guarantees full coverage of agent behavior. Improvements in these metrics consistently correlate with higher end-to-end task completion. We therefore use them as diagnostic signals for procedural fidelity rather than universal success criteria.

\subsubsection{\textbf{Output quality}} This metric evaluates the semantic, structural, and contextual consistency between the agent’s response and the reference answer.

To achieve scalable, fine-grained evaluation beyond lexical overlap, we adopt an LLM-based assessment framework using models such as Qwen-3 and DeepSeek as intelligent evaluators. This approach overcomes the limits of traditional metrics (e.g., BLEU, ROUGE) by jointly considering the response, reference, and contextual metadata (user query, prompt, and tool descriptions).

The evaluator scores five dimensions, denoted as \textbf{CRCFF metrics}:

\begin{itemize}[noitemsep, topsep=0pt, leftmargin=*]
\item \textbf{Correctness}: Whether the response accurately reflects factual content without errors or distortions.
\item \textbf{Relevance}: Whether it directly addresses user intent and aligns with task context.
\item \textbf{Completeness}: Whether it covers all required information without omissions.
\item \textbf{Fluency}: Whether the text is coherent, natural, and grammatically correct.
\item \textbf{Faithfulness}: Whether it remains logically consistent and free from hallucinated details.
\end{itemize}

Each is rated from 0–10, enabling interpretable, multidimensional evaluation of response quality.

\begin{table}[t]

\caption{Comparison of APIGen and Jenius-bench.}
\label{tab:Comparison_APIGen_JeniusBench_core}
\vskip 0.15in
\begin{center}
\begin{scriptsize}
\begin{sc}
\begin{tabularx}{0.5\textwidth}{@{}lXX@{}}
\toprule
Aspect & APIGen & Jenius-bench \\
\midrule
Dialogue Turn & Single & Multi \\
Data Source & LLM-generated & Human + Automatic \\
Verification & Format / Execution / Semantics & Trace + Review \\
Task Type & Simple calls & Chained reasoning \\
Tool Info & Basic meta & Contextual MCP \\
Memory & None & Cross-turn \\
Realism & Abstracted & Real-world \\
Scale & 60K / 21 cats. & 850 / 38 cats. \\
Focus & Call accuracy & Reasoning / Planning \\
\bottomrule
\end{tabularx}
\end{sc}
\end{scriptsize}
\end{center}
\vskip -0.1in
\end{table}

\subsubsection{\textbf{Token consumption}} Measures the computational input and output overhead of the agent by quantifying the number of input tokens fed to the LLM and generate output tokens. 

This metric captures the cumulative context length, including user queries, system prompts, tool descriptions, and historical dialogue states, and serves as an indicator of inference cost, memory usage, and scalability. 

\subsection{Benchmark Task Design and Dataset Construction}
To comprehensively evaluate the capabilities of autonomous agents in diverse operational contexts, we conducted experiments on two complementary benchmark datasets: one widely adopted public benchmark and one newly curated, real-world task-oriented dataset. These datasets differ significantly in their dialogue structure, tool complexity, metadata richness, and annotation rigor, enabling a multi-faceted assessment of agent performance across controlled and realistic scenarios.

\subsubsection{\textbf{APIGen: A Single-Turn Tool-Use Benchmark}} APIGen~\cite{liu2024apigen} evaluates models’ ability to interpret natural language queries into structured API calls. It contains 60K annotated samples across 21 categories (3.6K APIs, 2.3 parameters per call). Each instance includes a user query, candidate APIs, and a ground-truth function call.  

In our evaluation, we sample 800 instances from APIGen to ensure computational tractability while preserving task diversity. However, a key limitation of APIGen is its restriction to single-turn interactions: each instance represents an isolated tool-calling decision without contextual dependencies or multi-step reasoning. Furthermore, the API descriptions are minimal and syntactic in nature, lacking rich semantic annotations or behavioral specifications. To amplify the challenge and better stress our retrieval-augmented architecture, we extend the original candidate tool list for each query by randomly injecting 100 irrelevant tools sampled from its API pool, increasing the average candidate count. This augmented setup intensifies the need for accurate semantic matching and efficient top-$M$ filtering, thereby highlighting the effectiveness of our dynamic tool selection mechanism under high-noise conditions. The noise-injected APIGen setting serves as a stress test for retrieval robustness under extreme tool-set expansion. It is not intended to represent realistic tool registries. Our primary conclusions are drawn from Jenius-bench, which reflects real-world tool usage patterns.

\subsubsection{\textbf{Jenius-bench: A Multi-Turn, Semantically Enriched Task Dialogue Dataset}}

To address the limitations of synthetic and single-turn benchmarks, we introduce Jenius-bench, a novel, task-oriented dialogue dataset specifically constructed to evaluate advanced agent systems in realistic, interactive environments. A detailed comparison between APIGen and Jenius-bench is provided in Table~\ref{tab:Comparison_APIGen_JeniusBench_core}. Comprising 850 high-quality, human-curated samples, Jenius-bench covers a broad range of application domains—including travel planning, air and train ticket booking, web search, weather inquiry, website generation, and academic paper retrieval—spanning 38 distinct tool categories.

Crucially, Jenius-bench supports multi-turn dialogues with complex state evolution, where each interaction step includes detailed execution traces: the invoked tool (implemented as a fully encapsulated MCP-compliant module), its input arguments, the actual output returned, and the agent’s subsequent response. Unlike APIGen’s shallow metadata, tools in Jenius-bench are associated with rich semantic descriptions, including purpose, preconditions, expected outcomes, and usage examples, enabling deeper intent understanding and more robust orchestration.

Moreover, all trajectories in Jenius-bench are derived from real user-agent interactions and undergo rigorous manual review and annotation by domain experts to ensure correctness, coherence, and faithfulness. The final responses and execution paths serve as gold-standard references, making the dataset particularly suitable for evaluating not only output quality but also procedural fidelity and error recovery in long-horizon tasks.

Jenius-bench is designed for reproducibility. We explicitly exclude time-sensitive or non-deterministic queries, such as date-specific weather or ticket availability. Each task is associated with a fixed tool set and deterministic execution constraints. Reference trajectories specify required tools and ordering, ensuring stable evaluation across runs.

\subsection{Comparative Agent Configuration and Ablation Design}
To systematically evaluate the individual and cumulative contributions of the proposed optimization modules, we conducted controlled ablation studies across four progressively enhanced agent configurations. Each variant shares a consistent architectural backbone and inference protocol, differing only in the inclusion of specific components, thereby enabling fine-grained attribution of performance changes to targeted improvements.

All agents are evaluated on both APIGen and Jenius-bench using our assessment framework: process fidelity is measured using the 4T metrics, while outcome quality is assessed via the CRCFF metrics. Additionally, total token consumption is reported to evaluate computational efficiency.

The evaluated agents are defined as follows:
\begin{itemize}[noitemsep, topsep=0pt, leftmargin=*]
\item \textbf{Base:} A standard ReAct-style agent implementing the observe–think–act loop, generating reasoning traces and tool calls dynamically. It serves as the canonical baseline for modular agent systems.
\item \textbf{B-P:} Incorporates adaptive prompt generation, dynamically reformulating system prompts based on task intent classification to improve contextual alignment.
\item \textbf{B-PT:} Further extends B-P with context-aware tool orchestration, retrieving relevant tools via dense semantic matching using Qwen3-Embedding.
\item \textbf{Jenius:} The complete system combining all three modules, adding hierarchical memory management with dialogue- and session-level summarization for stable long-horizon reasoning.

\end{itemize}

\begin{table}[t]
\caption{Execution Fidelity Results on APIGen}
\label{tab:task-behavior-apigen}
\vskip 0.15in
\begin{center}
\begin{small}
\begin{sc}
\begin{tabular}{lccccc}
\toprule
Model & TCR  & TFR & TIR & TPS \\
\midrule
Base & 0.8150 & 0.1800 & 0.0050 & 0.8150 \\
B-P & 0.8275 & 0.1675 & 0.0050 & 0.8275\\
B-PT & 0.8375 & 0.1587 & \textbf{0.0038} & 0.8375 \\
Jenius & \textbf{0.8500} & \textbf{0.1362} & 0.0138 & \textbf{0.8500} \\
\bottomrule
\end{tabular}
\end{sc}
\end{small}
\end{center}
\vskip -0.1in
\end{table}

\begin{table}[t]
\caption{Execution Fidelity Results on Jenius-bench}
\label{tab:task-behavior-Jenius-bench}
\vskip 0.15in
\begin{center}
\begin{small}
\begin{sc}
\begin{tabularx}{0.5\textwidth}{lcccc}
\toprule
Model & TCR & TFR & TIR & TPS \\
\midrule
Base          & 0.5659 & \textbf{0.0329} & 0.4012 & 0.5968 \\
B-P        & 0.7271\textcolor{green!50!black}{~($\uparrow$16\%)} & 0.0859 & 0.1871 & 0.7491 \\
B-PT     & 0.7494 & 0.0718 & 0.1788 & 0.7740 \\
Jenius        & \textbf{0.7647} & 0.0753 & \textbf{0.1600} & \textbf{0.7847} \\
\bottomrule
\end{tabularx}
\end{sc}
\end{small}
\end{center}
\vskip -0.1in
\end{table}

\subsection{Experimental Results and Comparative Analysis}

\subsubsection{\textbf{Execution fidelity - 4T testing performances}}

\begin{table*}[t]
\caption{Output Qualit Results on Jenius-bench Evaluated by Qwen and DeepSeek}
\label{tab:text-gen-results}
\vskip 0.15in
\begin{center}
\begin{small}
\begin{sc}
\begin{tabular}{lcccccc}
\toprule
Evaluator & Model & Correctness & Relevance & Completeness & Fluency & Faithfulness \\
\midrule
\multirow{4}{*}{Qwen} 
& Base & 0.6741 & 0.8951 & 0.7722 & 0.9294 & 0.7919 \\
& B-P & 0.7259\textcolor{green!50!black}{~($\uparrow$5.2\%)} & 0.9204 & 0.7873 & 0.9563 & 0.8390 \\
& B-PT & 0.7572 & 0.9380 & 0.8027 & 0.9763 & 0.8560 \\
& Jenius & \textbf{0.7580} & \textbf{0.9447} & \textbf{0.8088} & \textbf{0.9771} & \textbf{0.8766} \\
\midrule
\multirow{4}{*}{DeepSeek} 
& Base & 0.7890 & 0.9245 & 0.7898 & 0.9546 & 0.8291 \\
& B-P & 0.8310\textcolor{green!50!black}{~($\uparrow$4.2\%)} & 0.9357 & 0.8153 & 0.9613 & \textbf{0.8654} \\
& B-PT & 0.8330 & \textbf{0.9428} & \textbf{0.8174} & \textbf{0.9711} & 0.8652 \\
& Jenius & \textbf{0.8350} & 0.9400 & 0.8143 & 0.9686 & 0.8636 \\
\bottomrule
\end{tabular}
\end{sc}
\end{small}
\end{center}
\vskip -0.1in
\end{table*}

The results demonstrate distinct patterns between the two datasets.  
On the public APIGen benchmark (Table~\ref{tab:task-behavior-apigen}), all models perform similarly, with the \textit{Base} model already reaching a high task completion rate (TCR = 0.8150). Subsequent optimizations (B-P, B-PT, and Jenius) yield only marginal improvements up to 0.8500, reflecting that APIGen mainly assesses pattern matching within familiar tool templates rather than genuine reasoning or generalization.

In contrast, on the Jenius-bench dataset (Table~\ref{tab:task-behavior-Jenius-bench}), the benefits of modular optimization are substantial. B-P improves TCR from 0.5659 to 0.7271 and reduces redundant invocations (TIR) by 21\%. B-PT further enhances accuracy (TCR = 0.7494, TPS = 0.7740), while the fully integrated Jenius model achieves the best overall performance (TCR = 0.7647, TPS = 0.7847), exhibiting coherent multi-turn tool orchestration.
Overall, the saturation observed on APIGen can be attributed to extensive pre-training coverage, whereas Jenius-bench underscores the robustness and adaptability of our modular agent optimization framework.

\begin{figure}[t]
\centering
\includegraphics[width=0.5\textwidth]{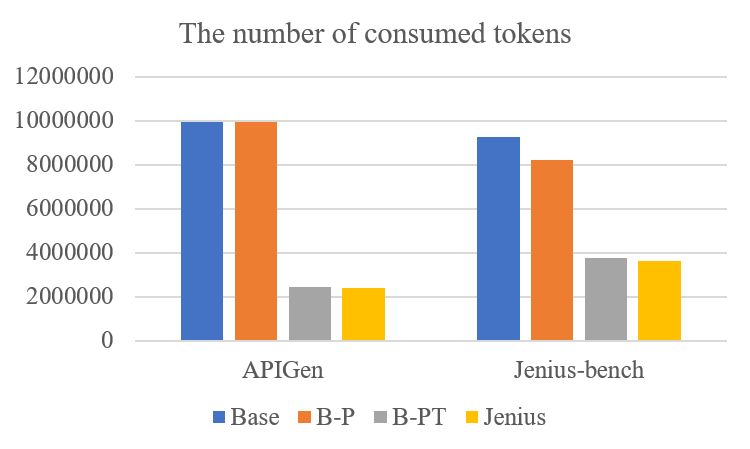}
\caption{Token consumption of evaluated agents on APIGen and Jenius-bench.}
\label{fig:token-consumption11}
\end{figure}

\subsubsection{\textbf{Output quality - CRCFF evaluation metrics}}

We evaluate generated responses using two independent LLM-based evaluators (Qwen and DeepSeek) across CRCFF. As shown in Table~\ref{tab:text-gen-results}, the evaluation is conducted solely on \textbf{Jenius-bench}, since APIGen employs virtual API stubs that lack real textual outputs for reliable quality analysis. Jenius-bench, in contrast, features realistic multi-turn dialogues where reasoning traces and tool results jointly determine response quality. In Jenius-bench, session-level summarization is triggered in approximately 32\% of tasks. Each task typically spans four to five user–agent turns, with one to two tool invocations per turn. When triggered, summarized segments are compressed to roughly 35–40\% of their original token length. This design treats summarization as a regular execution mechanism rather than an exceptional fallback. The results from both evaluators reveal consistent trends: the Adaptive Prompt Generation module significantly enhances relevance and fluency by producing more stable, context-aware prompts; the Context-Aware Tool Orchestration module improves correctness and completeness through precise semantic alignment and richer tool utilization; and Hierarchical Memory Management strengthens faithfulness by preserving contextual coherence in long-horizon tasks. Overall, the fully integrated \textbf{Jenius} achieves the best CRCFF performance, with an 8–10\% improvement in correctness and fluency over the baseline, demonstrating the complementary effectiveness of the three optimization modules in producing coherent, accurate, and faithful responses.


\subsubsection{\textbf{Token consumption analysis}}
We compare token efficiency of all agents on APIGen and Jenius-bench (Figure~\ref{fig:token-consumption11}), focusing on tokens used during tool invocations. In APIGen, only tool-call tokens are measured, as its virtual API stubs produce no real outputs.

Across both datasets, results show a clear downward trend. On APIGen, total tokens drop from 9.96M in \textit{Base} to 2.41M in \textit{B-PT}, with the full \textit{Jenius} model maintaining similar efficiency (2.46M). On Jenius-bench, usage decreases from 9.27M to 3.65M, a reduction of over 60\% compared with the baseline. These results highlight the cumulative efficiency gains brought by our modular design.

Specifically, Adaptive Prompt Generation reduces redundant reasoning, Context-Aware Tool Orchestration minimizes unnecessary tool calls, and Hierarchical Memory Management preserves coherence without increasing overhead. Overall, the consistent decline in token usage from Base to Jenius confirms that each module contributes complementary benefits, enabling efficient and interpretable agent behavior for real-world deployment.

\section{Real-World Deployment and Experience}
\label{sec:industrial}

This section presents the real-world deployment and operational insights of our anonymous agent system of Jenius in a production environment. We outline the deployment context, summarize key usage metrics over a four-month period, and distill practical lessons learned from operating the system at scale.

\subsection{Deployment Context}
\label{sec:deployment}

The Jenius system was initially rolled out through a gray-scale release on May 21, 2025, followed by full public deployment on May 25, 2025. Since then, the system has undergone continuous optimization guided by the methodology introduced in prior sections. 

Hosted on Alibaba Cloud, the architecture adopts a multi-node distributed design to ensure modularity, scalability, and fault isolation. Key components include: the Jenius Agent Engine for core reasoning and tool orchestration, several dedicated MCP Tool Servers for secure function execution, an external LLM Service for language generation, and Langfuse as the observability and logging infrastructure for tracing interactions and debugging. 

All services are containerized and orchestrated using Kubernetes with integrated CI/CD pipelines, enabling automated deployment and version management. Inter-service communication is secured via mutual TLS, ensuring data integrity and confidentiality. This modular design enables independent scaling and monitoring of each subsystem while maintaining end-to-end privacy and reliability.

\subsection{Scale and Usage Patterns}
\label{sec:scale}
\begin{itemize}[noitemsep, topsep=0pt, leftmargin=*]
    \item As of now, after six months of experience accumulation and system iteration, the system’s daily active user (DAU) count has increased to 265; in contrast, during the first month of public operation, the system achieved a daily active user (DAU) count of 42 and a monthly active user (MAU) count of 1,277, indicating early but concentrated adoption.
    \item User engagement was characterized by high interaction frequency: on average, each active user initiated 2.06 conversation sessions per day, with each session consisting of 2.4 turns (i.e., message exchanges between user and agent). Furthermore, every turn involved an average of 1.15 tool calls, reflecting the system’s reliance on external functions to fulfill complex requests.
    \item  Geographically, usage spanned 34 countries, with the majority of traffic originating from Mainland China (97.07\%), followed by the United States (1.26\%) and Japan and Malaysia (combined $<1$\%). Within China, users were distributed across nearly 300 cities, with the highest concentrations in Beijing (8.67\%), Suzhou (8.03\%), Shanghai (6.01\%), and Guangzhou (5.39\%). This distribution suggests strong initial traction in major technology and business hubs. 
\end{itemize}

Despite the relatively small user base, the depth of interaction demonstrates meaningful utility and sustained user engagement.

\subsection{Lessons Learned}
\label{sec:lesson}

Real-world deployment of the Jenius Agent provides critical insights into the design and optimization of LLM-based agents.

\begin{enumerate}[noitemsep, topsep=0pt, leftmargin=*]
    \item \textbf{Balancing Internal Knowledge and External Tool Use}: We observed that relying on tool calls for every query, especially those answerable from the model’s internal knowledge, leads to unnecessary computational overhead and increased latency. By strategically leveraging the LLM’s pretrained knowledge for common or factual queries (e.g., definitions, general knowledge), we significantly reduced token consumption and API costs while improving response speed. This hybrid approach enhances efficiency without compromising accuracy.
    
    \item \textbf{Prudent Tool Invocation with Semantic Fidelity}: A recurring issue was the generation of spurious tool calls: either invoking non-existent tools or fabricating parameters not grounded in user input. To mitigate this, we implemented stricter prompting strategies and post-generation validation checks to ensure that tool invocations are both necessary and semantically justified. This principle of “think before calling” is essential for maintaining system reliability and trustworthiness.

    \item \textbf{Explainability in Multi-Step Reasoning}: For complex tasks requiring several sequential tool calls, users often found the interaction confusing due to the lack of intermediate explanations. We found that providing concise, human-readable reasoning traces, such as summarizing the plan or justifying each step, greatly improves transparency and user comprehension. Explicit planning and explanation are therefore crucial for usability in long-horizon tasks.

    \item \textbf{Error Handling and Retry Behavior}: The default retry mechanism for failed tool calls led to excessive resource consumption and degraded user experience, particularly when errors were irrecoverable (e.g., network timeouts, invalid inputs). Blind retries not only wasted compute resources but also caused frustrating delays. We addressed this by introducing adaptive retry policies based on error types and user feedback signals, reducing redundant executions by over 60\% in subsequent iterations.

    \item \textbf{Specialized Handling for High-Cost or Fragile Tools}: Certain tools require domain-specific safeguards. For instance, the 'url\_reader' tool initially failed to extract URLs due to mismatched patterns in user input; this was resolved by incorporating robust regular expression matching. Similarly, long-running operations such as PPT or web application generation were prone to being triggered inappropriately (e.g., interpreting “generate a fitness plan” as a request for a PowerPoint). We introduced intent disambiguation layers and explicit confirmation prompts to prevent unintended activation of expensive or irreversible actions.
\end{enumerate}

These findings demonstrate that our closed–loop optimization framework is not only academically sound but also practically impactful, providing a blueprint for integrating LLM–based agents into large–scale industrial systems.

\section{Discussion and Conclusion}
\label{sec:discussion}
This work enhances autonomous agent performance through three complementary modules—adaptive prompt generation, context-aware tool orchestration, and hierarchical memory management. These modules jointly improve reasoning stability, tool invocation accuracy, and context retention, leading to consistent gains across text quality, execution fidelity, and token efficiency. The evaluation framework combines procedural, semantic, and efficiency metrics, revealing strong overall improvements while also highlighting limitations such as incomplete capture of hidden reasoning steps and the existence of multiple valid tool-use paths. Despite these challenges, large-scale deployment confirms the reliability and generalizability of the proposed framework.

Future work will focus on more flexible and outcome-oriented evaluation, accommodating diverse reasoning trajectories and tool sequences without penalizing valid outputs. Integrating user satisfaction, decision-making cost, and latency into evaluation will further align metrics with real-world performance. Operationally, adaptive strategies that reconfigure modules based on dynamic task contexts and multi-agent collaboration for distributed problem-solving represent key directions. The resulting system, Jenius, demonstrates stable performance improvements, efficient token utilization, and high user satisfaction in real-world use, providing both a validated empirical foundation and a practical blueprint for scalable, production-ready autonomous agent systems.





\nocite{langley00}

\bibliography{ref}

@inproceedings{yao2023react,
  title={React: Synergizing reasoning and acting in language models},
  author={Yao, Shunyu and Zhao, Jeffrey and Yu, Dian and Du, Nan and Shafran, Izhak and Narasimhan, Karthik and Cao, Yuan},
  booktitle={International Conference on Learning Representations (ICLR)},
  year={2023}
}

@article{xi2025rise,
  title={The rise and potential of large language model based agents: A survey},
  author={Xi, Zhiheng and Chen, Wenxiang and Guo, Xin and He, Wei and Ding, Yiwen and Hong, Boyang and Zhang, Ming and Wang, Junzhe and Jin, Senjie and Zhou, Enyu and others},
  journal={Science China Information Sciences},
  volume={68},
  number={2},
  pages={121101},
  year={2025},
  publisher={Springer}
}

@article{wang2024survey,
  title={A survey on large language model based autonomous agents},
  author={Wang, Lei and Ma, Chen and Feng, Xueyang and Zhang, Zeyu and Yang, Hao and Zhang, Jingsen and Chen, Zhiyuan and Tang, Jiakai and Chen, Xu and Lin, Yankai and others},
  journal={Frontiers of Computer Science},
  volume={18},
  number={6},
  pages={186345},
  year={2024},
  publisher={Springer}
}

@article{grattafiori2024llama,
  title={The Llama 3 Herd of Models},
  author={Dubey, Abhimanyu and Jauhri, Abhinav and Pandey, Abhinav and Kadian, Abhishek and Al-Dahle, Ahmad and Letman, Aiesha and Mathur, Akhil and Schelten, Alan and Yang, Amy and Fan, Angela and others},
  journal={CoRR},
  year={2024}
}

@article{masterman2024landscape,
  title={The landscape of emerging ai agent architectures for reasoning, planning, and tool calling: A survey},
  author={Masterman, Tula and Besen, Sandi and Sawtell, Mason and Chao, Alex},
  journal={arXiv preprint arXiv:2404.11584},
  year={2024}
}

@article{qin2024tool,
  title={Tool learning with foundation models},
  author={Qin, Yujia and Hu, Shengding and Lin, Yankai and Chen, Weize and Ding, Ning and Cui, Ganqu and Zeng, Zheni and Zhou, Xuanhe and Huang, Yufei and Xiao, Chaojun and others},
  journal={ACM Computing Surveys},
  volume={57},
  number={4},
  pages={1--40},
  year={2024},
  publisher={ACM New York, NY}
}

@article{singh2025agentic,
  title={Agentic retrieval-augmented generation: A survey on agentic rag},
  author={Singh, Aditi and Ehtesham, Abul and Kumar, Saket and Khoei, Tala Talaei},
  journal={arXiv preprint arXiv:2501.09136},
  year={2025}
}

@article{yang2023auto,
  title={Auto-gpt for online decision making: Benchmarks and additional opinions},
  author={Yang, Hui and Yue, Sifu and He, Yunzhong},
  journal={arXiv preprint arXiv:2306.02224},
  year={2023}
}

@inproceedings{topsakal2023creating,
  title={Creating large language model applications utilizing langchain: A primer on developing llm apps fast},
  author={Topsakal, Oguzhan and Akinci, Tahir Cetin},
  year={2023}
}

@article{nakajima2023task,
  title={Task-driven autonomous agent utilizing gpt-4, pinecone, and langchain for diverse applications},
  author={Nakajima, Yohei},
  journal={See https://yoheinakajima. com/task-driven-autonomous-agent-utilizing-gpt-4-pinecone-and-langchain-for-diverse-applications (accessed 18 April 2023)},
  volume={2},
  pages={3},
  year={2023}
}

@article{cemri2025multi,
  title={Why do multi-agent llm systems fail?},
  author={Cemri, Mert and Pan, Melissa Z and Yang, Shuyi and Agrawal, Lakshya A and Chopra, Bhavya and Tiwari, Rishabh and Keutzer, Kurt and Parameswaran, Aditya and Klein, Dan and Ramchandran, Kannan and others},
  journal={arXiv preprint arXiv:2503.13657},
  year={2025}
}

@article{e2025rag,
  title={From RAG to Multi-Agent Systems: A Survey of Modern Approaches in LLM Development},
  author={e Aquino, Gustavo de Aquino and de Azevedo, N{\'a}dila da Silva and Okimoto, Leandro Youiti Silva and Camelo, Leonardo Yuto Suzuki and de Souza Bragan{\c{c}}a, Hendrio Luis and Fernandes, Rubens and Printes, Andre and Cardoso, F{\'a}bio and Gomes, Raimundo and Torn{\'e}, Israel Gondres},
  year={2025},
  publisher={Preprints}
}

@inproceedings{wu2024autogen,
  title={Autogen: Enabling next-gen LLM applications via multi-agent conversations},
  author={Wu, Qingyun and Bansal, Gagan and Zhang, Jieyu and Wu, Yiran and Li, Beibin and Zhu, Erkang and Jiang, Li and Zhang, Xiaoyun and Zhang, Shaokun and Liu, Jiale and others},
  booktitle={First Conference on Language Modeling}
}

@article{schick2023toolformer,
  title={Toolformer: Language models can teach themselves to use tools},
  author={Schick, Timo and Dwivedi-Yu, Jane and Dess{\`\i}, Roberto and Raileanu, Roberta and Lomeli, Maria and Hambro, Eric and Zettlemoyer, Luke and Cancedda, Nicola and Scialom, Thomas},
  journal={Advances in Neural Information Processing Systems},
  volume={36},
  pages={68539--68551},
  year={2023}
}

@article{fei2025mcp,
  title={MCP-Zero: Proactive Toolchain Construction for LLM Agents from Scratch},
  author={Fei, Xiang and Zheng, Xiawu and Feng, Hao},
  journal={arXiv preprint arXiv:2506.01056},
  year={2025}
}

@article{zhou2025multi,
  title={Multi-Agent Design: Optimizing Agents with Better Prompts and Topologies},
  author={Zhou, Han and Wan, Xingchen and Sun, Ruoxi and Palangi, Hamid and Iqbal, Shariq and Vuli{\'c}, Ivan and Korhonen, Anna and Arik, Sercan O},
  year={2025}
}

@article{hou2025model,
  title={Model context protocol (mcp): Landscape, security threats, and future research directions},
  author={Hou, Xinyi and Zhao, Yanjie and Wang, Shenao and Wang, Haoyu},
  journal={arXiv preprint arXiv:2503.23278},
  year={2025}
}

@inproceedings{khattab2024dspy,
  title={Dspy: Compiling declarative language model calls into state-of-the-art pipelines},
  author={Khattab, Omar and Singhvi, Arnav and Maheshwari, Paridhi and Zhang, Zhiyuan and Santhanam, Keshav and Haq, Saiful and Sharma, Ashutosh and Joshi, Thomas T and Moazam, Hanna and Miller, Heather and others},
  booktitle={The Twelfth International Conference on Learning Representations}
}

@inproceedings{marvin2023prompt,
  title={Prompt engineering in large language models},
  author={Marvin, Ggaliwango and Hellen, Nakayiza and Jjingo, Daudi and Nakatumba-Nabende, Joyce},
  booktitle={International conference on data intelligence and cognitive informatics},
  pages={387--402},
  year={2023},
  organization={Springer}
}

@article{wang2025recursively,
  title={Recursively summarizing enables long-term dialogue memory in large language models},
  author={Wang, Qingyue and Fu, Yanhe and Cao, Yanan and Wang, Shuai and Tian, Zhiliang and Ding, Liang},
  journal={Neurocomputing},
  volume={639},
  pages={130193},
  year={2025},
  publisher={Elsevier}
}

@article{bensal2025reflect,
  title={Reflect, Retry, Reward: Self-Improving LLMs via Reinforcement Learning},
  author={Bensal, Shelly and Jamil, Umar and Bryant, Christopher and Russak, Melisa and Kamble, Kiran and Mozolevskyi, Dmytro and Ali, Muayad and AlShikh, Waseem},
  journal={arXiv preprint arXiv:2505.24726},
  year={2025}
}

@article{liu2025biomedtools,
  title={BioMedTools: a language model-powered community for biomedical computational tools},
  author={Liu, Sheng and Xing, Huadong and Han, Mengying and Zhang, Dachuan and Gong, Linlin and Liu, Dongliang and Chen, Junni and Cai, Pengli and Hu, Qian-Nan},
  journal={bioRxiv},
  pages={2025--05},
  year={2025},
  publisher={Cold Spring Harbor Laboratory}
}

@inproceedings{wu2024stateflow,
  title={StateFlow: Enhancing LLM Task-Solving through State-Driven Workflows},
  author={Wu, Yiran and Yue, Tianwei and Zhang, Shaokun and Wang, Chi and Wu, Qingyun},
  booktitle={NeurIPS 2024 Workshop on Open-World Agents}
}

@article{liu2024apigen,
  title={Apigen: Automated pipeline for generating verifiable and diverse function-calling datasets},
  author={Liu, Zuxin and Hoang, Thai and Zhang, Jianguo and Zhu, Ming and Lan, Tian and Tan, Juntao and Yao, Weiran and Liu, Zhiwei and Feng, Yihao and RN, Rithesh and others},
  journal={Advances in Neural Information Processing Systems},
  volume={37},
  pages={54463--54482},
  year={2024}
}

@misc{ibm2024acp,
  author       = {{IBM BeeAI}},
  title        = {Introduction to Agent Communication Protocol (ACP)},
  howpublished = {\url{https://docs.beeai.dev/acp/alpha/introduction}},
  year         = {2024},
  note         = {Accessed: April 2025}
}

@misc{google2024a2a,
  author       = {{Google}},
  title        = {{Agent2Agent (A2A) Protocol Documentation}},
  howpublished = {\url{https://google.github.io/A2A/}},
  year         = {2024},
  note         = {Accessed: April 2025}
}

@article{zhang2025qwen3,
  title={Qwen3 Embedding: Advancing Text Embedding and Reranking Through Foundation Models},
  author={Zhang, Yanzhao and Li, Mingxin and Long, Dingkun and Zhang, Xin and Lin, Huan and Yang, Baosong and Xie, Pengjun and Yang, An and Liu, Dayiheng and Lin, Junyang and others},
  journal={arXiv preprint arXiv:2506.05176},
  year={2025}
}

@inproceedings{yao2022react,
  title={React: Synergizing reasoning and acting in language models},
  author={Yao, Shunyu and Zhao, Jeffrey and Yu, Dian and Du, Nan and Shafran, Izhak and Narasimhan, Karthik R and Cao, Yuan},
  booktitle={The eleventh international conference on learning representations},
  year={2022}
}

@inproceedings{langley00,
 author    = {P. Langley},
 title     = {Crafting Papers on Machine Learning},
 year      = {2000},
 pages     = {1207--1216},
 editor    = {Pat Langley},
 booktitle     = {Proceedings of the 17th International Conference
              on Machine Learning (ICML 2000)},
 address   = {Stanford, CA},
 publisher = {Morgan Kaufmann}
}
\bibliographystyle{mlsys2025}


\end{document}